\documentclass[sigconf]{acmart}

\AtBeginDocument{%
  }

\usepackage[normalem]{ulem}
\usepackage{booktabs}
\usepackage{amsfonts}
\usepackage{graphicx}
\usepackage{hhline}
\usepackage{pifont}
\usepackage{listings}
\usepackage{float}
\usepackage{algpseudocode}
\usepackage{subfigure}
\usepackage{makecell}
\usepackage{multirow}
\usepackage{stmaryrd}
\usepackage{setspace}
\usepackage{xcolor}
\usepackage{array}
\usepackage{enumitem}
\usepackage{balance}

\newcommand{\cmark}{\ding{51}}%
\newcommand{\xmark}{\ding{55}}%
\newcommand{\ie}{\emph{i.e.,} }

\newcommand{\eg}{\emph{e.g.,} }

\setcopyright{cc}
\setcctype{by-nc-sa}
\copyrightyear{2025}
\acmYear{2025}
\acmConference[KDD '25]{Proceedings of the 31st ACM SIGKDD Conference on Knowledge Discovery and Data Mining V.2}{August 3--7, 2025}{Toronto, ON, Canada}
\acmBooktitle{Proceedings of the 31st ACM SIGKDD Conference on Knowledge Discovery and Data Mining V.2 (KDD '25), August 3--7, 2025, Toronto, ON, Canada}
\acmDOI{10.1145/3711896.3737174}
\acmISBN{979-8-4007-1454-2/2025/08}

\begin{document}

\title{Unsupervised Time Series Anomaly Prediction with Importance-based Generative Contrastive Learning}

\author{Kai Zhao}
\orcid{0000-0002-5159-2312}
\affiliation{
  \institution{Aalborg University}  
  \city{Aalborg}
  \country{Denmark}
}
\email{kaiz@cs.aau.dk}

\author{Zhihao Zhuang}
\orcid{0009-0004-0748-4754}
\affiliation{%
  \institution{East China Normal University}
  \city{Shanghai}
  \country{China}
  }
\email{zhuangzhihao@stu.ecnu.edu.cn}

\author{Chenjuan Guo}
\authornote{Corresponding Author}
\orcid{0000-0002-4516-4637}
\affiliation{
  \institution{East China Normal University}  
  \city{Shanghai}
  \country{China}
}
\email{cjguo@dase.ecnu.edu.cn}

\author{Hao Miao}
\orcid{0000-0001-9346-7133}
\affiliation{
  \institution{Aalborg University}  
  \city{Aalborg}
  \country{Denmark}
}
\email{haom@cs.aau.dk}


\author{Christian S. Jensen}
\orcid{0000-0002-9697-7670}
\affiliation{%
  \institution{Aalborg University}
  \city{Aalborg}
  \country{Denmark}
  }
\email{csj@cs.aau.dk}

\author{Yunyao Cheng}
\orcid{0000-0002-1819-4056}
\affiliation{
  \institution{Aalborg University}  
  \city{Aalborg}
  \country{Denmark}
}
\email{yunyaoc@cs.aau.dk}

\author{Bin Yang}
\orcid{0000-0002-1658-1079}
\affiliation{
  \institution{East China Normal University}
  \city{Shanghai}
  \country{China}
}
\email{byang@dase.ecnu.edu.cn}

\renewcommand{\shortauthors}{Kai Zhao et al.}

\begin{abstract} 
We study the problem of time series anomaly prediction, which is relevant to a range of real-world applications. Existing anomaly prediction methods rely on labeled training data for achieving acceptable accuracy. However, such data may be difficult to obtain; and in real-time deployments, anomalies can occur that were not seen in labeled data, thus making them difficult to predict. We provide a theoretical analysis and propose an Importance-based Generative Contrastive Learning method (IGCL) for unsupervised anomaly prediction. IGCL employs a controlled diffusion module to produce anomaly precursor patterns. Next, ICGL learns contextual representations to extract temporal dependencies from pairs of normal time series and anomaly precursors. IGCL is then able to predict anomalies by identifying anomaly precursors that will evolve into future anomalies. To address challenges caused by potentially complex precursor combinations involving multiple variables, we propose a memory bank with importance scores that stores representative samples adaptively and generates more complex anomaly precursors. Extensive experiments on nine benchmark datasets offer evidence that the proposed method is able to outperform state-of-the-art baselines.
\end{abstract}




\maketitle

\newcommand\kddavailabilityurl{https://doi.org//10.5281/zenodo.15561219}

\ifdefempty{\kddavailabilityurl}{}{
\begingroup\small\noindent\raggedright\textbf{KDD Availability Link:}\\
The source code of this paper has been made publicly available at \url{\kddavailabilityurl}.
\endgroup
}

\section{Introduction}

\begin{figure*}[t]
\subfigure[Supervised anomaly prediction learned with labeled anomaly precursors.]{
\label{fig:Illustration}
\includegraphics[width=0.5\linewidth]{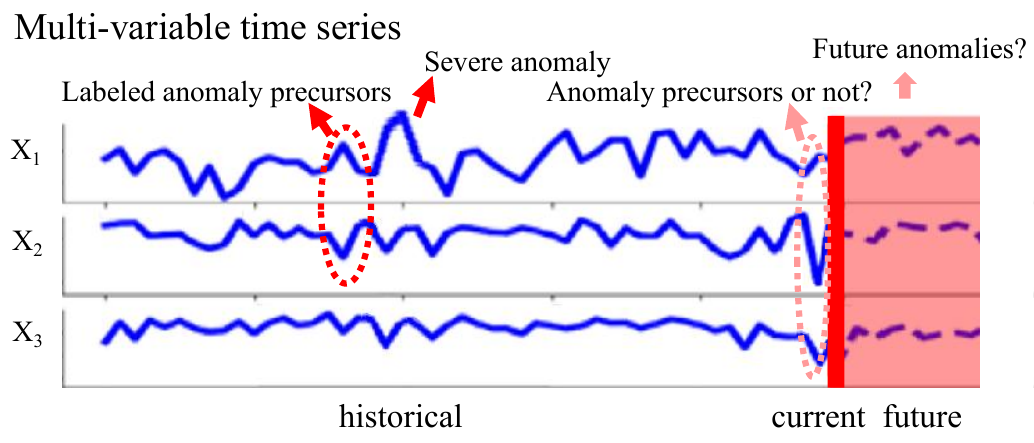}
}
\subfigure[Different anomaly precursor combinations in multi-variable time series.]{
\label{fig:differentanomaly}
\includegraphics[width=0.46\linewidth]{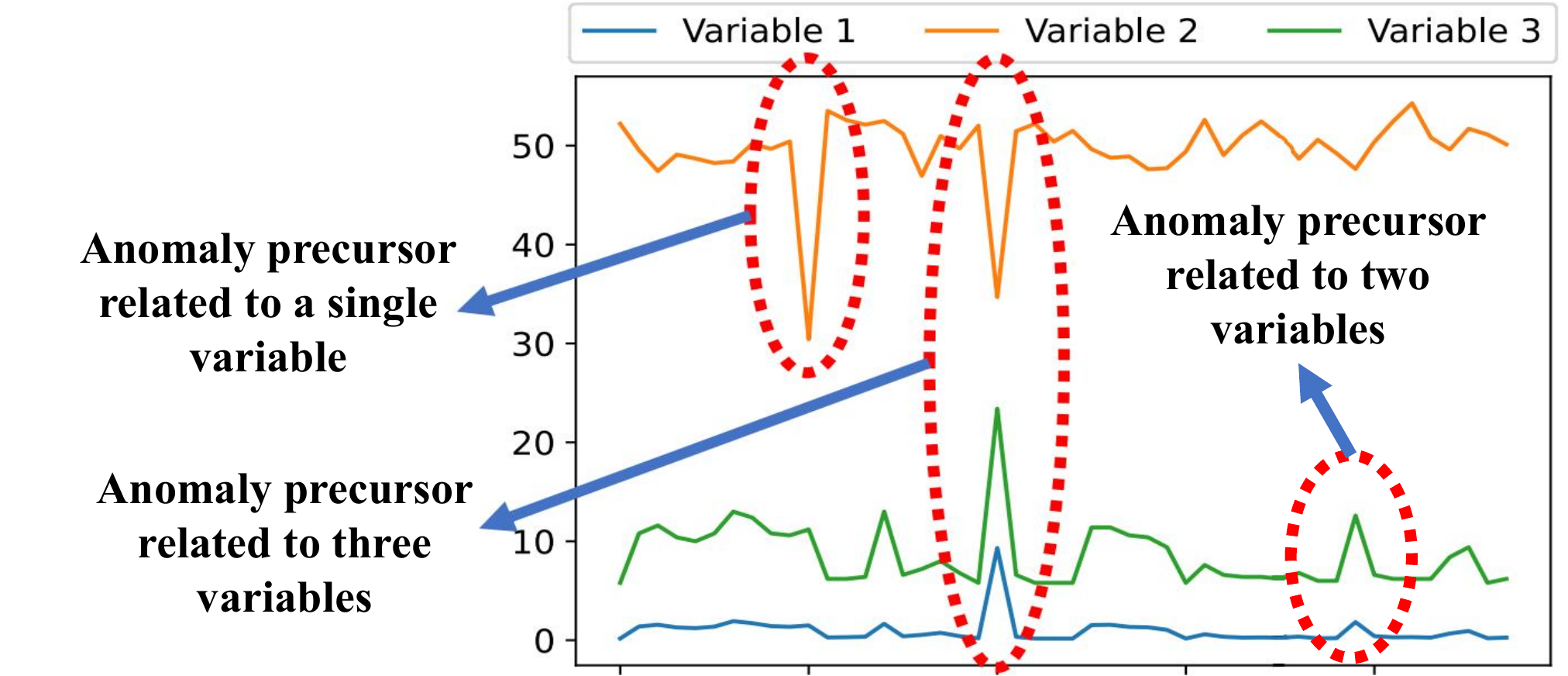}
}
\caption{Our motivation.}
\label{fig:example}
\end{figure*}

Much data collected by real-world applications can be modeled as time-dependent observations that form multi-variable time series~\cite{yinArchitectureSearch, outyin2, yinmiao, yunyaovldb24, kaiyin}. For example, computing systems record runtime indicators~\cite{PSM, SMD-omni} and environmental systems record water quality~\cite{NIPS-DS, yinGraphAttentionRecurrent}. 
Based on the current time series data, anomaly prediction makes real-time judgments on whether anomaly signals start to occur that indicate that severe anomalies will occur after the current time~\cite{Precursor-of-Anomaly, outpre1, outpre00, outpre0}, as exemplified in Figure~\ref{fig:Illustration}. Anomaly prediction plays an important role in scenarios that preserve some form of safety and prioritize losses caused by anomalies~\cite{outpre2, EarlyAnomaly, EarlyAnomalynew, Earthquake}, \eg in health care~\cite{outpre1} or drinking-water systems~\cite{Precursor-of-Anomaly}.
For example, anomaly prediction can help address impending pollution or faults in advance~\cite{outpre00, outpre0} or can warn of extreme environmental conditions~\cite{Earthquake}.

Only a few classification-based studies~\cite{EarlyAnomaly, EarlyAnomalynew, Earthquake, outpre2} exist related to the time series anomaly prediction. Existing methods predict future anomalies by determining whether current observations contain anomaly signals that are the start of deviations from normal behaviors~\cite{outpre1, outpre00, outpre0}. 
Such early signals are called the anomaly \textit{precursor}s~\cite{Precursor-of-Anomaly}. However, to obtain accurate results, existing methods heavily rely on supervised learning, where anomaly precursors are given during training. 

Yet, in many real-world applications, there is insufficient labeled training data to enable supervised learning because manual labeling is costly~\cite{outreview}.
Further, unexpected anomalies~\cite{anomlyreview1, yinnewout} that 
did not appear prior to occurring during real-time deployment, and thus were not considered during training and make supervised methods fail in practice~\cite{anomlyreview}. 
Therefore, unsupervised time series anomaly prediction is of high interest. However, this has not been studied in depth nor analyzed theoretically due to its challenging nature, as outlined next.

{\bf Challenge 1:} In the setting of unsupervised anomaly prediction, there is no labeled anomaly precursor data to enable models to learn temporal dependencies.
Although PAD~\cite{Precursor-of-Anomaly} proposes to use unsupervised anomaly detection methods~\cite{CAE-Ensemble, Autoencoder, PUAD} with a specific threshold to identify the precursors, their performance is limited when compared to supervised learning methods~\cite{outpre0}. Without labeled data, existing methods cannot learn temporal dependency features from pairs of a successive normal sub-sequence and precursor, where the data just begins to deviate from normal~\cite{outpre1}, which is crucial to enable anomaly prediction~\cite{Earthquake, outpre00}. As a result, existing unsupervised methods do not perform well in anomaly prediction. 

{\bf Challenge 2:} We are facing potentially very complex anomaly precursor combinations in multi-variable time series, which cause very high time and space complexity. As shown in Figure~\ref{fig:differentanomaly}, for any $n$-variable time series, there can be as many as $O(2^n)$ potential anomaly precursor combinations~\cite{outpre0}, where different anomaly precursors may appear in each individual variable or in variable combinations~\cite{HINkai, Searchkai}. 
It is very challenging to learn temporal dependencies comprehensively from this many anomaly precursor combinations.

We propose a novel anomaly prediction method called \textbf{IGCL}, which is short for \textbf{I}mportance-based \textbf{G}enerative \textbf{C}ontrastive \textbf{L}earn-ing for unsupervised time series anomaly prediction. 
To address {\bf Challenge 1}, we employ a generative contrastive learning architecture. We propose an anomaly precursor pattern generation module that employs a diffusion-based transformation and variance regularization, to generate different kinds of potential precursor patterns labeled as negative data using Gaussian noise. We also propose an overlapping window-based temporal convolutional network (TCN) architecture to efficiently extract temporal dependencies and learn positive and negative contextual representations for contrastive learning. More specifically, the model distinguishes normal sub-sequences from anomaly precursors, where pairs of successive normal sub-sequences are positive samples and pairs of a normal sub-sequence and an anomaly precursor, which contain temporal changes from normal to abnormal, are negative samples.

To address {\bf Challenge 2}, we propose a memory bank with importance-based scores to adaptively store representative anomaly precursors, which helps generate more complex anomaly precursor combinations related to different variables. First, anomaly precursor patterns can be generated efficiently as only one variable is considered during each iteration.
Second, the memory bank, with its fixed size $K$ that could be far smaller than $O(2^n)$, is initialized with generated anomaly precursors related to one variable. Then, other one-variable anomaly precursor patterns are generated iteratively and are injected into the previously generated anomaly precursors stored in the memory bank, thereby accumulating complex anomaly precursor combinations involving more variables. The memory bank only stores representative anomaly precursors and drops anomaly precursors that are already distinguished. We summarize our contributions as follows.

\begin{itemize}[leftmargin=3.5mm]
     \item We propose a generative contrastive learning architecture with a diffusion-based transformation and variance regularization to generate different precursor patterns and learn temporal dependencies to distinguish between normal and anomaly precursors. 
     \item We propose a memory bank with importance scores to store representative negative samples, which helps to generate more complex negative samples.
     \item Experiments on nine benchmarks from different domains show our method outperforms the state-of-the-art baselines.
\end{itemize}

\section{PRELIMINARIES} 
\label{sec:preliminary}

\subsection{Definitions}
\label{subsec:Formalization}

We first formalize the unsupervised time series anomaly prediction problem. Frequently used notation is summarized in Table~\ref{tab:Notations}.

\begin{table}[t]
\caption{{Notation}}
\label{tab:Notations}
\resizebox{0.49\textwidth}{!}
{
\begin{tabular}{ l | p{215pt} } \hline
Notation & Explanation \\\hline \hline
$\mathbf{x}_{t}$ & The observed data at timestamp $t$ from the $N$-variable time series $\mathbf{X}$\\\hline
$\mathbf{X}_{t+1:t+f}$ & \makecell[l]{The sub-sequence of the time series $\mathbf{X}$ from time\\ $t+1$ to $t+f$, with sequence length of $f$} \\\hline
$\mathbf{X}^i_{t-h:t}$ & The data of the $i$-th variable starting from $t-h$ to $t$\\\hline
$\mathbb{X}^b_{T, h}$ & \makecell[l]{The set of most recent $b$ sub-sequences with length\\  $h+1$ before the current timestamp $T$\\ $\mathbb{X}^b_{T, h}=\{ \mathbf{X}_{T-h:T}, \cdots, \mathbf{X}_{T-b+1-h:T-b+1}\}$ }
\\\hline
${N}(\mu,\sigma^2)$ & \makecell[l]{The multi-variable Gaussian distribution with mean \\ vector $\mu$ and  covariance matrix $\sigma^2$}  \\\hline 
$\mathbf{R}^i_{t-h:t}$ & The anomaly precursor patterns for the $i$-th variable\\\hline
$S$ & The max steps for diffusion-based transformation \\\hline
$\mathbf{v}_{t}$ & The feature vector extracted from time series at timestamp $t$\\\hline
$z_{t}^{+}$ & \makecell[l]{The representations from $(\mathbf{X}_{t-2h-1:t-h-1},\mathbf{X}_{t-h:t})$ to \\estimate $\mathcal{N} = P (\cdot | \mathbf{H_{t-h}}, \mathbf{H_t} \text{  is normal})$}\\ \hline
$z_{t}^{-}$ & \makecell[l]{The representations from $(\mathbf{X}_{t-2h-1:t-h-1},\mathbf{X}^-_{t-h:t})$ to \\estimate $\mathcal{A} = P (\cdot | \mathbf{H_{t-h}}, \mathbf{H_t} \text{  is an anomaly precursor})$}\\ \hline
$K$ & The max size of the memory bank\\\hline
\hline
\end{tabular}
}
\end{table}

\noindent \textbf{A Multi-variable Time Series} is represented as $\mathbf{X} = \langle \mathbf{x}_{1}, \mathbf{x}_{2}, \cdots, \mathbf{x}_{T} \rangle$ $\in\mathbb{R}^{N\times T}$, where $N$ is the number of variables collected at each timestamp, $T$ is the total number of historical timestamps, $\mathbf{x}_{t}\in\mathbb{R}^{N}$ denotes the $N$-dimensional observations at timestamp $t$. We use $\mathbf{X}_{t+1:t+f}\in\mathbb{R}^{N\times f}$ to indicate the data during the time window $[t+1, t+f]$, use $\mathbf{X}_{t+1:t+f}^{i}\in\mathbb{R}^{f}$ to indicate the data of the $i$-th variable, and use $\mathbf{X}_{t}^{i}\in\mathbb{R}^{1}$ to indicate the data of the $i$-th variable at timestamp $t$. We also use $T$ to denote the current timestamp.

\noindent \textbf{A Sub-sequence} $\mathbf{X}_{t_1:t_2}$ is a continuous subset of $\mathbf{X}$ from $t_1$ to $t_2$. The set of $b$ most recent sub-sequences before the current timestamp $T$, each with sequence length of $h+1$, is denoted as $\mathbb{X}^b_{T, h} = \{\mathbf{X}_{T-h:T}, \mathbf{X}_{T-1-h:T-1}, \cdots, \mathbf{X}_{T-b+1-h:T-b+1}\}$.

\noindent \textbf{An Anomaly Precursor} is a sub-sequence $\mathbf{X}_{T-h:T}$ containing early signals that just start deviating from normal and followed by future anomalies in $\mathbf{X}_{T+1:T+f}$~\cite{outpre00, outpre0, Precursor-of-Anomaly}, where $h$ and $f$ are hyper-parameters for the look-back and look-forward windows.

\noindent \textbf{Unsupervised time series anomaly prediction.}
At any current timestamp $T$, given the historical observations $\mathbf{X}$, where we use a set of $b$ most recent sub-sequences $\mathbb{X}^b_{T, h} = \{\mathbf{X}_{T-h:T}, \mathbf{X}_{T-1-h:T-1},$ $\cdots, \mathbf{X}_{T-b+1-h:T-b+1}\}$, the goal is to determine a real-time anomaly score $\hat{p}_{T}$ to indicate whether there are anomaly precursors currently and more anomalies in $\mathbf{X}_{T+1: T+f}$. The transformation of anomaly scores into binary labels is done by applying a threshold $\delta$~\cite{Precursor-of-Anomaly}, \ie if $\hat{p}_{T} < \delta$, the future sub-sequence $\mathbf{X}_{T+1: T+f}$ is normal; and if $\hat{p}_{T} \ge \delta$, the future sub-sequence $\mathbf{X}_{t+1: t+f}$ is abnormal. 
Thus, we formulate the time series anomaly prediction problem as  that of finding a model with a mapping function $\mathcal{F}$ and $\hat{p}_{T}=\mathcal{F}_\theta(\mathbb{X}^b_{T, h}),$
where $\theta$ is the parameters of the model $\mathcal{F}$.
For the unsupervised time series anomaly prediction problem, we do not have the true score ${p}_{T}$ as there is no labeled data.

\subsection{Theoretical Analysis}
\label{subsec:theoretical}

We analyze unsupervised time series anomaly prediction using Maximum Mean Discrepancy (MMD) theory~\cite{MMD}. 
We utilize Markov Chains~\cite{Markov}, widely used in time series analytics~\cite{Markov2}, 
to model sequence-based probabilities. We use random variables $\mathbf{H_{t-h-1}}$ and $\mathbf{H_t}$ to denote what observations the sub-sequences $\mathbf{X}_{t-2h-1:t-h-1}$ and $\mathbf{X}_{t-h:t}$ may be observed during two successive time-windows $[t-2h-1, t-h-1]$ and $[t-h, t]$, respectively.  Then, we define 

\ \ \ \ \ \ \ \ \ \ \ \ 
 $\mathcal{N} = P (\cdot | \mathbf{H_{t-h-1}}=\mathbf{X}_{t-2h-1:t-h-1}, \mathbf{H_t} \text{  is normal} )$

\noindent to denote the Markov conditional distribution of the normal sub-sequence observed during time-window $[t-h, t]$, given that its previous observations during time-window $[t-2h-1, t-h-1]$ are $\mathbf{X}_{t-2h-1:t-h-1}$. Similarly, we define 

\noindent  \,\,\,\, \,\,  $\mathcal{A} = P (\cdot |\mathbf{H_{t-h-1}}= \mathbf{X}_{t-2h-1:t-h-1}, \mathbf{H_t} \text{ is anomaly precursor} )$

\noindent to denote the Markov conditional distribution of the anomaly precursor observed during time-window $[t-h, t]$. We present the theorem here and its proof in Appendix~\ref{Appendix:Proof}.

\noindent \textbf{Theorem: } Under the unsupervised setting, the future anomaly is unpredictable if we cannot identify its anomaly precursor in the current observation that follows the same distribution as the normal previous sub-sequences do.
The future anomaly is predictable if we can identify its anomaly precursor that follows a distribution different from the normal previous sub-sequences.  $\square$

Thus, our model aims to distinguish normal distribution $\mathcal{N} = P (\cdot | \mathbf{H_{t-h-1}}, \mathbf{H_t}  \text{ is normal} )$ from any potential different distributions $\mathcal{A} = P (\cdot | \mathbf{H_{t-h-1}},$ $\mathbf{H_t} \text{ is anomaly}$  $\text{precursor})$.

\begin{figure*}[t]
\centering
\includegraphics[width=0.98\textwidth]{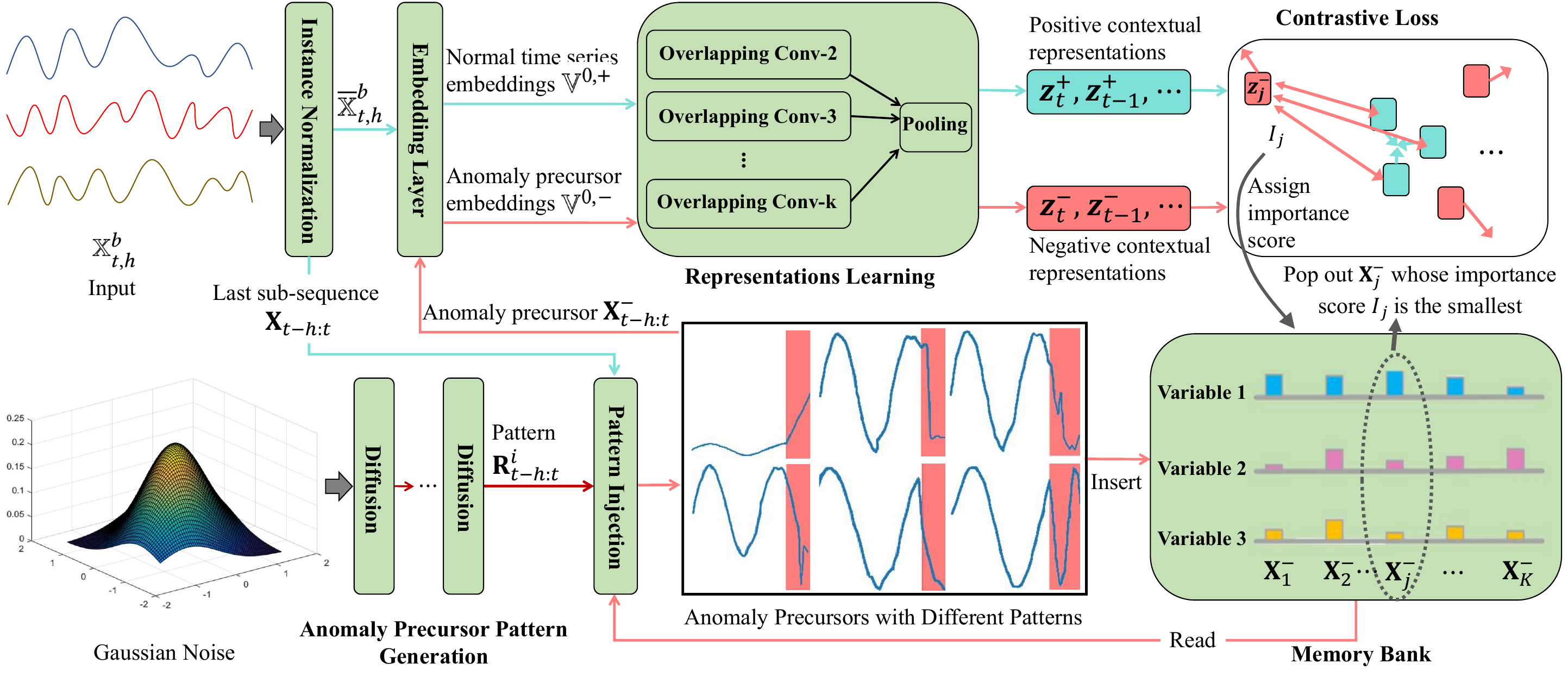}
\caption{The overall architecture.
}
\label{fig:framework}
\end{figure*}

\section{Methodology} 
\label{sec:methodology}

We present the overall architecture in Figure~\ref{fig:framework}, which consists of anomaly precursor pattern generation, positive and negative representations learning, contrastive loss, and the memory bank.

For any timestamp $t$, the model takes as input the set of $b$ available historical sub-sequences $\mathbb{X}^b_{t, h} = \{\mathbf{X}_{t-h:t}, \mathbf{X}_{t-1-h:t-1},\cdots\cdots, $ \\$\mathbf{X}_{t-b+1-h:t-b+1}\}$. We first apply Instance Normalization~\cite{InstanceNorm} on the observation of each time series variable, \ie $\mathbf{X}^i_{t'}$ in the set $\mathbb{X}^b_{t, h}$, and output $\overline{\mathbb{X}}^b_{t, h}$, to learn stable components~\cite{RevIn} from time series.

Embedding layer takes as input the normalized $\overline{\mathbb{X}}^b_{t, h}$, maps the observations at each timestamp $t'$, \ie $\mathbf{X}_{t'} \in\mathbb{R}^{N}$, to a $d$-dimensional vectors  $\mathbf{v}_{t'} \in\mathbb{R}^{d}$, which extracts the dense feature from normal time series, and outputs a set of feature vectors, \ie $\mathbb{V}^{0, +}$.

\textbf{Anomaly precursor pattern generation} takes as input a randomly generated Gaussian noise $x \sim {N}(0,I)$, and iteratively outputs any potential anomaly precursor pattern, \ie $\mathbf{R}^i_{t-h:t}$, for a randomly selected $i$-th time series variable. 

Pattern injection module injects $\mathbf{R}^i_{t-h:t}$ into the last sub-sequence $\mathbf{X}_{t-h:t}$ in the selected $i$-th time series variable, and outputs a time series anomaly precursor, \ie $\mathbf{X}^-_{t-h:t} = \mathbf{X}^i_{t-h:t} + \mathbf{R}^i_{t-h:t}$, which is fed into the embedding layer to get precursor feature vectors $\mathbb{V}^{0, -}$.

\textbf{Positive and negative representations learning} takes as input all possible pairs of successive sub-sequences from $\mathbb{V}^{0, +}$ and $\mathbb{V}^{0, -}$. For example, this module takes as input a pair of normal sub-sequences $\mathbf{X}_{t-2h-1:t-h-1}$ and $\mathbf{X}_{t-h:t}$ from $\mathbb{V}^{0, +}$, and outputs a positive contextual representation $z^+_t$ that learns their normal temporal dependencies. This module also takes as input a pair of a normal sub-sequence $\mathbf{X}_{t-2h-1:t-h-1}$ from $\mathbb{V}^{0, +}$ and an anomaly precursor $\mathbf{X}^-_{t-h:t}$ from $\mathbb{V}^{0, -}$, and outputs a negative contextual representation $z^-_t$ that learns their abnormal temporal dependencies. 

With \textbf{contrastive loss}, the model distinguishes $\mathcal{N} = P (\cdot |$ $\mathbf{H_{t-h-1}},$ $\mathbf{H_t}$ is normal$)$ and $\mathcal{A} = P (\cdot | \mathbf{H_{t-h-1}}, \mathbf{H_t}$ is anomaly  $\text{precursor})$ by pushing $z^-_t$ away from $z^+_t$. Intuitively, it clusters positive representations, \eg $z^+_t$, that belong to $\mathcal{N}$ together and dissociates negative representations, \eg $z^-_t$, that belong to $\mathcal{A}$ away.

{\bf The memory bank} stores the representative anomaly precursors. The generated anomaly precursor patterns are iteratively injected into existing anomaly precursors in the memory bank, to accumulate more complex anomaly precursor combinations. 
The importance score $I_j$ for the anomaly precursor $\mathbf{X}^{-}_j$ is calculated by its similarity with the normal samples. Finally, the memory bank pops out the anomaly precursor with the smallest importance score, which the model can easily distinguish from normal already.

\subsection{Anomaly Precursor Pattern Generation}
\label{subsec:anomalyDistribution}

As experimental results show diffusion models outperform other generative models for time series~\cite{DiffAD, D3R}, we propose diffusion-based transformation to generate diverse kinds of potential anomaly precursors from a simple Gaussian distribution, with our variance regularization to simulate precursor patterns that start deviating from normal.

The denoising diffusion model consists of two processes, \ie the pollution process with $S$ steps and its reverse denoising diffusion process, where $S$ is the max diffusion steps. Specifically, we use $x^0= \mathbf{R}^i_{t-h:t} \in\mathbb{R}^{h}$ to denote any potential anomaly precursor pattern that could be more complicated than Gaussian patterns. The pollution process is shown in Appendix~\ref{Appendix:noise pollution}. Our anomaly precursor pattern generation module is based on denoising diffusion process $p_{\theta}(x^{s-1}|x^s)$, reversing random Gaussian noise $x^S$ to produce more complicated potential anomaly precursor patterns: 
\begin{equation}
{x}^{s-1} \sim p_{\theta}(x^{s-1}|x^s) = \frac{1}{\sqrt{1-\beta^{s}}}\left({x}^{s} - \sqrt{\beta^{s}} \varepsilon_\theta ({x}^{s}) \right),
\end{equation}
\noindent where $\varepsilon_\theta ({x}^{s}) \sim {N}(\mu_{s},{\sigma_{s}}^2)$, and $\mu_{s}$ is the mean value parameter and $\sigma_{s}$ is the variance parameter to be learned. We use Multi-layer Perceptron (MLP) layers to model with $\mu^{s}$ and $\sigma^{s}$:

\;\;$\mu^{s}, \sigma^{s}  = MLP \big (concat({x}^{s}, s, \mathbf{X}^i_{t-h:t})\big),\  \forall s\in \{S, S-1, \cdots, 1 \},$

\noindent where $x^S$ is a random Gaussian noise, $x^s$ is the denoised state after each step, and $i$ denotes this process aiming to generate anomaly precursor patterns for the $i$-th variable. We have $\sigma^{s}\varepsilon + \mu^{s} \sim  {N}(\mu_{s},{\sigma_{s}}^2)$,
where $\varepsilon \sim {N}(0,I)$, and $\sigma^{s}\varepsilon + \mu^{s}$ is derivable with respective to $\mu^{s}$ and $\sigma^{s}$ which are the outputs from the MLP layers. Thus, the final denoising diffusion process is 
\begin{equation}
{x}^{s-1} = \frac{1}{\sqrt{1-\beta^{s}}}\left({x}^{s} - \sqrt{\beta^{s}} (\sigma^{s}\varepsilon +  \mu^{s})  \right),
\end{equation}
\noindent where we randomly sample the Gaussian noise $x^S$, and apply the denoising diffusion process step by step to get the anomaly precursor pattern $\mathbf{R}^i_{t-h:t} = x^0$.

We further propose regularization loss on parameters ${N}(\mu_{s}, {\sigma_{s}}^2)$, where the variance ${\sigma_{s}}^2$ controls the diversity of generated anomaly precursor patterns, and ${\sigma_{s}}^2$ and the mean value $\mu_{s}$ controls the degree of deviation from normal. 
First, the variance ${\sigma_{s}}^2$ should not be too large, as a large variance will produce severe anomalies rather than anomaly precursors with starting deviation from normal. Second, ${\sigma_{s}}^2$ should not be too small, as a small variance will generate less diverse~\cite{DiffAD} precursor patterns. Therefore, we propose to use the Kullback-Leibler divergence as regularization to make ${N}(\mu_{s}, {\sigma_{s}}^2)$ be close to ${N}(\mu_{s}, I)$ where ${\sigma_{s}}^2$ should be close to the unit normal variance $I$ which is neither too large nor too small~\cite{DBLP:journals/corr/vae}. For any ${N}(\mu, {\sigma}^2)$, we have 
\begin{equation}
KL\Big({N}(\mu,{\sigma}^2)\Big\Vert N(\mu,I)\Big) = \frac{1}{2}\Big(-\log \sigma^2+\sigma^2-1\Big). 
\end{equation}
The proof is in Appendix~\ref{Appendix:Regularization}. Thus, our analytical solution of the regularization loss is:
\begin{equation}
\label{eq:regularization}
\begin{aligned}
   \mathcal{L}_{r} = \sum_{s = 1}^{S} \frac{1}{2}\Big(-\log \sigma_{s}^2+\sigma_{s}^2-1\Big)
\end{aligned}
\end{equation}

\begin{figure*}[t]
\centering
\includegraphics[width=0.7\textwidth]{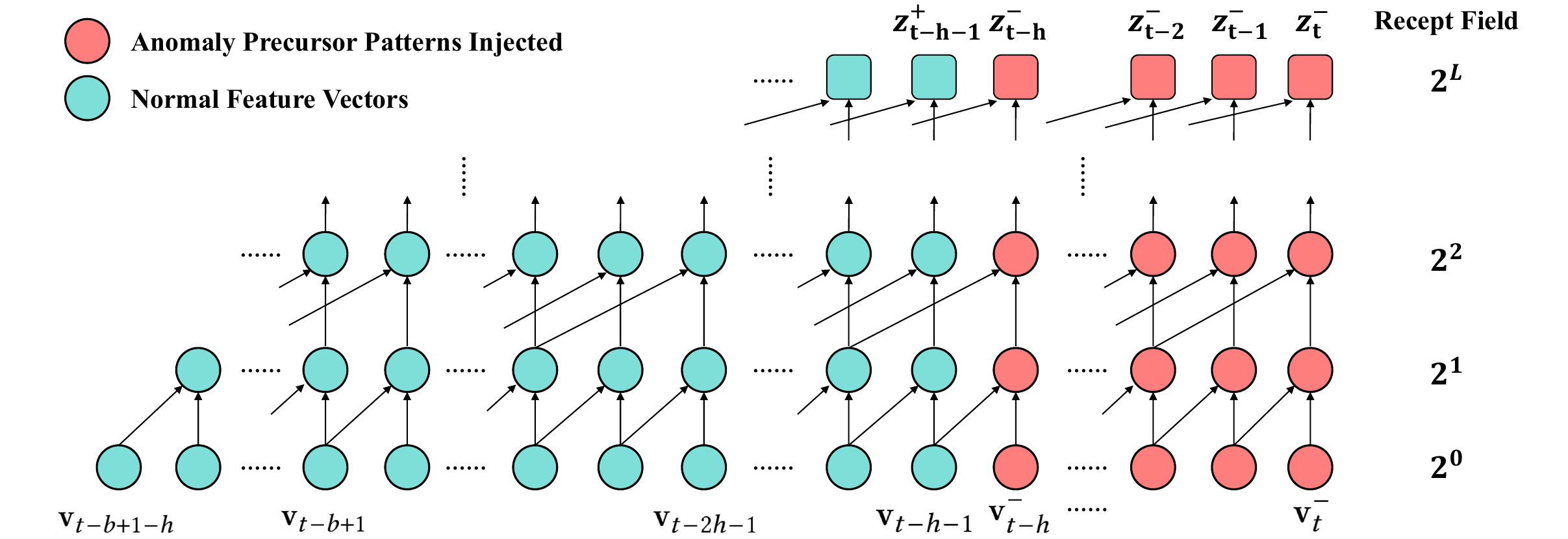}
\caption{The overlapping window-based temporal convolutional network with a kernel size $k$ of 2.
}
\label{fig:TCN}
\end{figure*}

\subsection{Positive and Negative Representations}
\label{subsec:TCN}

We propose an overlapping window-based TCN to learn contextual representations from the all possible pairs of successive sub-sequences followed by the Markov Chain and extract the normal and abnormal temporal dependencies at once. 

To be more specific, our embedding layer firstly maps the normalized time series data, \ie $\mathbf{X}_{t'} \in\mathbb{R}^{N}$ with $t' \in [t-b+1-h, t]$, equivalent to the set $\mathbb{X}^b_{t, h}$, into the $d$-dimensional feature vectors $\mathbf{v}_{t'} \in\mathbb{R}^{d}$, which aim to extract the dense features from time series for each timestamp. Further, the input to the embedding layer can also be coupled with other auxiliary attributes, such as the encoding of timestamps. The auxiliary attributes at timestamp $t'$ are represented as $\mathbf{a}_{t'} \in\mathbb{R}^{F}$, where $F$ is the total dimensions of the auxiliary attributes. Thus, the embedding layer concatenates the original time series data $\mathbf{X}_{t'}$ and the auxiliary attributes
\begin{equation}
\mathbf{f}_{t'}=concat(\mathbf{X}_{t'}, \mathbf{a}_{t'}) \; \forall  t' \in [t-b+1-h, t],
\end{equation}
as its input, and maps it to the feature vector by:
\begin{equation}
\mathbf{v}_{t'}=\sigma(W_e\, \mathbf{f}_{t'}), \; \forall  t' \in [t-b+1-h, t],
\end{equation}
where $\mathbf{v}_{t'}\in\mathbb{R}^{d}$ is the learned feature vector, $\sigma$ is an activation function, and $W_e\in\mathbb{R}^{d\times (N+F)}$ is the learnable neural network parameters which extract the feature from the time series data and the auxiliary attributes at each timestamp. Following the same way, we can also get the feature vector $\mathbf{v}^-_{t'}\in\mathbb{R}^{d}$, where $ t'\in [t-h, t]$, from the generated anomaly precursor $\mathbf{X}^-_{t-h:t}$.

Then, our overlapping window-based TCN uses the feature vectors $\mathbf{v}_{t'}$ and $\mathbf{v}^-_{t'}$ to learn the temporal dependencies for all possible pairs of successive sub-sequences at once. 
The temporal convolution operation slides over timestamps by skipping feature vectors with a certain dilation factor in different layers, as illustrated in Figure~\ref{fig:TCN}. 
More specifically, this module takes as input all the feature vectors, \ie $\mathbf{v}_{t'} $ with $t' \in [t-b+1-h, t]$ and $\mathbf{v}^-_{t'} $ with $t' \in [t-h, t]$: 
\begin{equation}
\label{eq:TCN}
\begin{aligned}
\mathbb{V}^{0,+} &= \big \langle \mathbf{v}_{t-b+1-h}, \cdots, \mathbf{v}_{t-h}, \cdots, \mathbf{v}_{t}\big\rangle,\\
\mathbb{V}^{0,-} &= \big \langle\mathbf{v}_{t-b+1-h}, \cdots, \mathbf{v}^-_{t-h}, \cdots, \mathbf{v}^-_{t}\big \rangle,
\end{aligned}
\end{equation}
and uses the learnable convolution $filter \in \mathbb{R}^{k}$ with the kernel size of $k$ to extract features and learn the temporal dependencies by:
\begin{equation}
\begin{aligned}
\mathbb{V}^{l,+}(t')  = \sum_{i=0}^{k-1} filter(i) \mathbb{V}^{l-1,+}(t'-k^{l-1}\times i), \, \text{for } 1 \le l \le L, \\
\mathbb{V}^{l,-}(t')  = \sum_{i=0}^{k-1} filter(i) \mathbb{V}^{l-1,-}(t'-k^{l-1}\times i), \, \text{for } 1 \le l \le L,
\end{aligned}
\end{equation}
where $k^{l-1}$ is the dilation factor in the layer $l-1$, $\mathbb{V}^{l,+}(t')$ and $\mathbb{V}^{l,-}(t')$ is the learned dependency features in the layer $l$ for timestamp $t'$, and $L \sim O(log(h))$ is the max number of TCN layers whose recept field is large than the whole window size of the pair of successive sub-sequences. We use different kernel sizes to extract features in parallel and pool them together, to capture different kinds of temporal dependencies with different kernel scales. Finally, it outputs all contextual representations for all pairs of successive sub-sequences at once, \eg $z^+_t = \mathbb{V}^{L,+}(t)$, $z^+_{t-1} = \mathbb{V}^{L,+}(t-1)$, $z^-_t = \mathbb{V}^{L,-}(t)$ and $z^-_{t-1} = \mathbb{V}^{L,-}(t-1)$. 
In this way, our total complexity is reduced to $O\big((b+h)log(h)\big)$, rather than the complexity of all pairs of sub-sequences as $O\big((b+h)^2\big)$. 

\subsection{The Objective Function}
\label{subsec:Loss}

We propose a contrastive strategy to distinguish $\mathcal{N}$ and $\mathcal{A}$ by pushing $z^-_t$ away from $z^+_t$. Intuitively, it clusters positive representations, \eg $z^+_t$, that belong to $\mathcal{N}$ together and dissociates negative representations, \eg $z^-_t$, that belong to $\mathcal{A}$ away. 
Each anchor $z^+_{t}$ is the contextual representation from the pair of the successive normal sub-sequences before timestamp $t$, which represents for $\mathcal{N} = P (\cdot | \mathbf{H_{t-h}}, \mathbf{H_t} \text{  is normal} )$. Thus, we make the anchor $z^+_{t}$ similar to the positive representations $z^+_{t-j}$ with $1\le j \le P$, which are from the most nearby successive normal sub-sequences before timestamp $t$ and they follow the normal data distribution. In order to avoid data leakage and be consistent with the inference stage, for each anchor $z^+_{t}$ the positive samples are always from the previous timestamps.
Meanwhile, the anchor $z^+_{t}$ should be different from the negative representation $z^-_{t}$, which learns the temporal dependencies from normal to the generated anomaly precursors and represents for anomaly precursors distribution $\mathcal{A}$. Thus, our contrastive loss function is: 

\begin{equation}
\begin{aligned}
\mathcal{L}_{c} = \sum_{i = t}^{t-h} -log 
\frac{\sum_{j = 1}^{P} exp\big(Sim(z^+_{i},z^+_{i-j})/\tau \big)}{exp\big(Sim(z^+_{i},z^-_{i})/\tau \big) + \sum_{j = 1}^{P} exp\big(Sim(z^+_{i},z^+_{i-j})/\tau \big)},
\end{aligned}
\end{equation}
where $\tau$ is the temperature parameter to control the softmax strength, $P$ is the hyper-parameter for the number of positive samples used, and $Sim$ is the similarity measurement, \eg Cosine Similarity, to distinguish whether the contextual representations belong to the same distributions. 

Finally, together with the regularization loss, we train the model using Adam gradient descent~\cite{Adam}:
\begin{equation}
\label{eq:loss}
\begin{aligned}
   \mathcal{L} = \mathcal{L}_{c} + \lambda \mathcal{L}_{r},
\end{aligned}
\end{equation}
where $\lambda$ is to control the strength of our diffusion regularization.

\begin{table*}[t]
\centering
\caption{Details of benchmark datasets.}
\label{table:BenchmarkDatasets}
\begin{tabular}{lllllll} \toprule
Dataset & Domain & $N$ \ & \makecell[l]{\#Training \ \\ (Unlabeled)} & \makecell[l]{\#Validation \ \\ (Unlabeled)} & \makecell[l]{\#Test \\ (Labeled)} \  & Anomaly Rate(\%) \\ \hline\hline
SMD & server machine & 38 & 566,724 & 141,681 & 708,420 & 4.16 \\
PSM & application server & 25 & 105,885 & 26,398 &  87,841 & 27.8 \\
MSL & NASA space sensor & 55 & 46,653 & 11,663 &  73,729 & 10.72 \\
SMAP & NASA space sensor & 25 & 108,146 & 27,036 & 427,617 & 13.13 \\
SWAN & space solar weather & 38 & 48,000 & 12,000 & 60,000 & 32.6 \\ 
SWaT & water treatment & 51 & 396,000 & 99,000 & 449,919 & 12.14 \\
GECCO & water quality & 9 & 55,408 & 13,852 & 69,261 & 1.1 \\ 
UCR & various domains & 1 & 1,790,679  & 447,670 & 6,143,541 & 0.6 \\ 
Credit & finance & 29 & 159,491 & 39,873 & 85,443 & 0.172 \\ 
\bottomrule
\end{tabular}
\end{table*}

\subsection{Memory Bank}

We propose to use a memory bank to store previously generated anomaly precursors and combine them with the current anomaly precursor pattern $\mathbf{R}^i_{t-h:t}$, to accumulate more complex anomaly precursor combinations related to more than one variable. 

To be more specific, the memory bank $M$ has a fixed size of $K$ and empirically $K \ll O(2^n)$ in our experiments:
\begin{equation}
\begin{aligned}
\mathbf{M} = \big\{\mathbf{X}^{-}_1, \mathbf{X}^{-}_2,\cdots, \mathbf{X}^{-}_K \big\},
\end{aligned}
\end{equation}
where each $\mathbf{X}^{-}_j$ denotes a stored anomaly precursor. 
The currently generated anomaly precursor pattern $\mathbf{R}^i_{t-h:t}$ which is only related to the $i$-th variable, is also injected into the anomaly precursors in the memory bank:
\begin{equation}
\begin{aligned}
\mathbf{X}^{-}_j \leftarrow \mathbf{X}^{-}_j + \mathbf{R}^i_{t-h:t}, 
\end{aligned}
\end{equation}
and thus we can get anomaly precursors that involve more than one variable. Then, similar to Equation~(\ref{eq:TCN}), we can get more negative contextual representations, \eg $z^{-}_{t,j}$ and $z^{-}_{t-1,j}$, from the anomaly precursor $\mathbf{X}^{-}_j$. The contrastive loss $\mathcal{L}_{c}$ is updated to
\begin{equation}
\begin{aligned}
\sum_{i = t}^{t-h} -log 
\frac{\sum_{j = 1}^{P} exp\big(Sim(z^+_{i},z^+_{i-j})/\tau \big)}{\sum_{j = 0}^{K} exp\big(Sim(z^+_{i},z^-_{i,j})/\tau \big) + \sum_{j = 1}^{P} exp\big(Sim(z^+_{i},z^+_{i-j})/\tau \big)},
\end{aligned}
\end{equation}
where $z^{-}_{i,0} = z^{-}_{i}$ is the negative contextual representations from the current anomaly precursor $\mathbf{X}^{-}_{t-h:t}$, and $z^{-}_{i,j}$ is the negative contextual representations from the $j$-th anomaly precursor $\mathbf{X}^{-}_j$ stored in the memory bank. 

Finally, the current anomaly precursor $\mathbf{X}^{-}_{t-h:t}$ is inserted into the memory bank which will have a size of $K+1$, \ie $\mathbf{X}^{-}_0 = \mathbf{X}^{-}_{t-h:t}$. Instead of using a first-in-first-out strategy to maintain the memory bank with a fixed size of $K$, we propose to pop out the anomaly precursor based on an importance score. The intuition is that the memory bank should store the hard and important negative samples for further model training, \ie the anomaly precursors that are not yet distinguished by the model, and should pop out not important samples, \ie the anomaly precursors that are already dissociated away from normal. The importance score $I_j$ for the anomaly precursor $\mathbf{X}^{-}_j$ is calculated by:
\begin{equation}
\begin{aligned}
I_j = \sum_{i = t}^{t-h} Sim(z^+_{i},z^-_{i,j}).
\end{aligned}
\end{equation}
Thus, we pop out the anomaly precursor with the smallest importance score after each iteration of model optimization.

\subsection{Inference}
\label{subsec:inference}

For each real-time $T$ in the inference stage, we output the probability score $\hat{p}_{T}$ to predict how likely there is an anomaly precursor currently and will be more future anomalies in $\mathbf{X}_{T+1, T+f}$:
\begin{equation}
\hat{p}_{T} = \sum_{j = 1}^{K} Sim(z^+_{T},z^-_{T,j}) - \sum_{j = 1}^{P} Sim(z^+_{T},z^+_{T-j}),
\end{equation}
where $K$ is the size of the memory bank with anomaly precursors as negative samples, and $P$ is the number of positive samples.

\begin{table*}
\centering
\caption{Overall performance. We report Precision, Recall, and F1-score results here, presented in percentages. }
\label{table:performance}
\resizebox{1.05\textwidth}{!}
{
\begin{tabular}{c|ccc|ccc|ccc|ccc|ccc|ccc|ccc} \hline
Dataset & \multicolumn{3}{c|}{PSM} & \multicolumn{3}{c|}{SMAP} & \multicolumn{3}{c|}{SWAN} & \multicolumn{3}{c|}{SWaT} & \multicolumn{3}{c|}{GECCO} & \multicolumn{3}{c|}{SMD} & \multicolumn{3}{c}{MSL}\\ \hline
Metric & P & R & F1 & P & R & F1 & P & R & F1 & P & R & F1 & P & R & F1 & P & R & F1 & P & R & F1\\ \hline\hline
DAGMM & 39.27 & 36.14 & 37.64 & 11.91 & 22.00 & 15.51 & 52.73 & 39.46 & 45.14 & 82.14 & 61.72 & 70.48 & 47.50 & 27.64 & 34.95 & 14.74 & 15.37 & 15.05 & 13.94 & 17.04 & 15.33\\
IForest & 39.39 & 38.82 & 39.10 & 11.79 & 18.67 & 14.45 & 53.85 & 39.07 & 45.28 & 75.39 & 62.22 & 68.17 & 42.75 & 29.89 & 35.18 & 15.25 & 18.28 & 16.63 & 13.82 & 17.11 & 15.29\\ \hline
K-means & 38.89 & 39.04 & {38.96} & 13.05 & 21.37 & 16.20 & 54.04 & 42.30 & 47.45 & 82.13 & 62.50 & 70.98 & 47.94 & 29.18 & 36.28 & 15.77 & 17.40 & 16.54 & 14.23 & 15.99 & 15.06\\
Deep-SVDD & 36.63 & 36.30 & {36.46} & 11.77 & 17.09 & 13.94 & 52.39 & 39.18 & 44.83 & 81.29 & 61.95 & 70.31 & 46.09 & 28.45 & 35.18 & 14.99 & 15.78 & 15.37 & 12.94 & 16.04 & 14.32 \\
THOC & 37.64 & 38.93 & 38.27 & 12.61 & 21.89 & 15.99 & 54.93 & 40.02 & 46.30 & 82.42 & 62.45 & 71.06 & 46.41 & 29.49 & 36.09 & 16.90 & 16.79 & 16.84 & 14.37 & 16.22 & 15.24\\ \hline
DCdetector & 28.97 & 12.05 & 17.02 & 8.04 & 11.30 & 9.40 & 21.04 & 10.94 & 14.40 & 46.45 & 36.36 & 40.79 & 1.76 & 3.46 & 2.33 & 3.71 & 9.03 & 5.26 & 2.35 & 3.46 & 2.80\\
PAD & 31.23 & 33.05 & 32.11 & 11.01 & 22.37 & 14.76 & 50.13 & 37.61 & 42.98 & 74.83 & 59.09 & 66.04 & 44.38 & 28.17 & 34.46 & 11.23 & 18.76 & 14.05 & 14.37 & 13.22 & 13.77\\\hline
ATransformer & 30.56 & 34.64 & 32.47 & 11.40 & 22.80 & 15.20 & 52.20 & 38.75 & 44.48 & 75.00 & 60.50 & 66.97 & 43.62 & 27.92 & 34.05 & 10.52 & 19.42 & 13.65 & 14.62 & 12.42 & 13.43\\
LSTM-VAE & 39.20 & 40.31 & 40.02 & 11.68 & 22.65 & 15.41 & 54.42 & 40.88 & 46.69 & 66.76 & 68.36 & 67.55 & 46.67 & 30.92 & 37.20& 14.21 & 16.42 & 15.24 & 16.18 & 15.28 & 15.72  \\
Omni & 39.45 & 40.88 & 40.15 & 11.90 & 23.39 & 15.77 & {54.83} & 41.00 & {46.92} & {85.73} & {58.57} & {69.59} & 50.55 & 29.65 & {37.38}& 16.22 & 18.19 & 17.15 & 13.66 & 21.66 & 16.75  \\ 
GANomaly & 37.02 & 43.06 & 39.81 & 12.18 & 23.41 & 16.02 & {54.44} & 40.91 & {46.72} & {83.99} & {59.77} & {69.84} & 50.06 & 29.23 & {36.91}& 15.06 & 21.39 & 17.68 & 15.36 & 17.30 & 16.27  \\ 
CAE-Ensemble & 40.18 & 41.99 & 41.07 & 15.88 & 26.68 & \underline{19.91} & {55.98} & 41.52 & {47.68} & {88.61} & {59.90} & \underline{71.48} & 51.67 & 29.92 & {37.90}  & 18.41 & 19.51 & \underline{18.94} & 20.35 & 18.27 & 19.26 \\ 
PUAD & 39.91 & 42.38 & 41.11 & 15.20 & 27.20 & 19.50 & {54.91} & 41.19 & {47.07} & {85.68} & {58.05} & {69.21} & 50.76 & 29.58 & {37.38}  & 16.37 & 21.85 & {18.72} & 20.04& 17.60 & 18.74\\ 
D3R & 40.73 & 42.21 & \underline{41.46} & 15.73 & 26.89 & 19.85 & {55.32} & 42.64 & \underline{48.16} & 84.13 & {61.85} & {71.29} & 51.24 & 30.91 & \underline{38.56}& 15.78 & 19.33 & 17.38 & 19.99 & 20.32 & \underline{20.15}  \\ \hline
\textbf{IGCL}  & 42.31 & 46.97 & \textbf{44.52} & 17.48 & 28.02 & \textbf{21.53} & 59.20 & 44.94 & \textbf{51.09} & 84.98 & 63.54 & \textbf{72.71} & 52.10 & 32.77 & \textbf{40.23} & 17.39 & 25.90 & \textbf{20.81} & 21.97 & 22.81 & \textbf{22.38}\\ \hline
\end{tabular}
}
\end{table*}

\section{EXPERIMENTS}
\label{sec:experiments}

\subsection{Experimental Settings}

\subsubsection{Datasets.}
Following existing works~\cite{Precursor-of-Anomaly, CAE-Ensemble, outreview2}, we validate the performance of our method with nine real-world benchmark datasets: SMD~\cite{SMD-omni}, PSM~\cite{PSM}, MSL~\cite{SMAP-MSL}, SMAP~\cite{SMAP-MSL}, SWAN~\cite{NIPS-DS}, SWaT~\cite{AnomalyTrans}, GECCO~\cite{NIPS-DS}, UCR~\cite{outreview2} and Credit~\cite{anomlyreview}.  More details of the datasets can be seen in Appendix~\ref{Appendix:Datasets} and Table~\ref{table:BenchmarkDatasets}.

\subsubsection{Baselines.}
We compare the unsupervised anomaly prediction method PAD~\cite{Precursor-of-Anomaly}. We also modify more state-of-the-art unsupervised anomaly detection methods into anomaly prediction following PAD:
\begin{itemize}[leftmargin=3.5mm]
\item Density-based: DAGMM~\cite{DAGMM} and IForest~\cite{KmeanIsolationForest}.
\item Clustering-based: K-means~\cite{outreview}, Deep-SVDD~\cite{DeepOneClass} and THOC~\cite{THOC}.
\item Contrastive-based: DCdetector~\cite{DC} and PAD~\cite{Precursor-of-Anomaly}.
\item Autoregression and reconstruction-based: LSTM-VAE~\cite{VQRAE}, ATra-nsformer~\cite{AnomalyTrans},  Omni~\cite{SMD-omni}, GANomaly~\cite{GANomaly}, CAE-Ensemble~\cite{CAE-Ensemble}, PUAD~\cite{PUAD} and D3R~\cite{D3R}.
\end{itemize}

We introduce the details of baselines in Section~\ref{sec:related}. More details of settings are in Appendix~\ref{Appendix:Settings}.

\subsubsection{Evaluation Metrics.}
Following the evaluation methodology in existing works~\cite{CAE-Ensemble, Precursor-of-Anomaly, anomlyreview}, we validated the performance of time series anomaly prediction models with Precision (P), Recall (R), F1-score (F1), and ROC-AUC~\cite{outyin1VUS}, where larger values indicate higher model performance and ROC-AUC enables evaluation without choosing the threshold. 
We do not use the point adjustment (PA) metrics, as many recent works have demonstrated that PA can lead to faulty performance evaluations~\cite{AffPR, D3R, AFF1, anomlyreview, outyin1VUS}. PA uses true labels to adjust the outputs of models, and it is known that using PA can result in state-of-the-art performance even with random guess~\cite{AffPR, AFF1}.

\subsection{Overall Comparison and Analysis}

Table~\ref{table:performance} and Table~\ref{table:AUCperformance} in Appendix~\ref{Appendix:More experimental results} show the overall anomaly prediction performance of different models on all datasets. We randomly repeat each method 3 times and report the average result, where the best F1-score and AUC-ROC are highlighted in bold and significantly outperform the underlined second-best ones.

Key observations are followed. First, IGCL consistently outperforms the state-of-the-art baseline methods on all datasets. It demonstrates that IGCL is able to learn the normal temporal dependencies and identify the abnormal temporal dependencies that change from the normal to the abnormal with our generated anomaly precursors, which finally improves the accuracy of anomaly prediction.  

Second, we observe that the contrastive-based, density-based, and clustering-based methods are unsatisfactory on most datasets. This is because they cannot generate anomaly precursors as negative samples to learn the abnormal temporal dependencies that change from normal to abnormal. They only learn with the non-labeled data as positive samples. Thus, their accuracy is limited as anomaly prediction requires learning both normal patterns and abnormal temporal dependencies.

Third, IGCL achieves the best accuracy compared to the autoregression and reconstruction-based methods, as they only learn to reconstruct each normal time series sub-sequence within each time window. They cannot explicitly learn the abnormal temporal dependencies from normal to anomaly precursors and thus have limited accuracy.
The experiments demonstrate that learning the temporal dependencies from a pair of successive time series sub-sequences that change from the normal to the anomaly precursor is crucial to the anomaly prediction.

\begin{table}[t]
\caption{Parameter sensitivity}
\label{tab:Accuracyparemeter}
\resizebox{0.44\textwidth}{!}
{
\begin{tabular}{c|ccc|ccc}\hline
Dataset               & \multicolumn{3}{c|}{SWaT} & \multicolumn{3}{c}{MSL} \\\hline
$K$               & P    & R   & F1 & P    & R   & F1     \\\hline
4    & 84.67  & 61.93  & 71.54  & 20.01  & 20.17  & 20.09   \\
8  & 84.23  & 62.48  & 71.74  & 20.38  & 21.42  & 20.89   \\
12  & 84.52  & 62.97  & 72.17  & 20.50  & 21.51  & 20.99   \\
16   & 84.21 & 63.19 & {72.20}  & 20.43 & 21.69 & {21.04}    \\
20   & 84.09  & 63.25  & 72.19  & 20.33  & 21.65  & 20.97     \\
24   & 84.98 & 63.54 & \textbf{72.71} & 21.97 & 22.81 & \textbf{22.38} \\
\hline
\end{tabular}}
\end{table}

\begin{table}[t]
\caption{Larger look-forward windows}
\label{tab:Largerlook-forward}
\resizebox{0.48\textwidth}{!}
{
\begin{tabular}{c|cc|cc|cc}\hline
Dataset               & \multicolumn{6}{c}{SWaT} \\\hline
$f$               & \multicolumn{2}{c|}{4}    & \multicolumn{2}{c|}{8}  & \multicolumn{2}{c}{12} \\\hline
Metric    & F1 & AUC  & F1 & AUC  &F1 & AUC  \\\hline
CAE-Ensemble & 71.48 & 81.71 & 68.62 & 74.92 & 62.70 & 73.32\\
D3R & 71.29 & 80.62 & 68.51 & 75.94 & 62.43 & 71.76\\
IGCL & \textbf{72.71} & \textbf{82.39} & \textbf{70.02} & \textbf{77.75} & \textbf{64.56} & \textbf{74.88}\\\hline
\end{tabular}
}
\end{table}

\subsection{Paremeter Sensitivity}
We evaluate the impact of $K$, the size of the memory bank. The experimental results are shown in Table~\ref{tab:Accuracyparemeter}.  
If we set a bigger size for the memory bank to store more representative anomaly precursors, our model will have better results. 
When $K$ increases to 24, our results are significantly better than the baselines, and $K=24$ is far smaller than $O(2^n)$, where $n$, \ie the total number of variables, is $51$ and $55$ for SWaT and MSL datasets, respectively.

\begin{table*}[t]
\caption{Parameter sensitivity}
\label{tab:look-backwindow}
\centering
\resizebox{0.92\textwidth}{!}
{
\begin{tabular}{c|cccc|cccc|cccc}\hline
Dataset               & \multicolumn{4}{c|}{SWaT} & \multicolumn{4}{c|}{PSM} & \multicolumn{4}{c}{SWAN} \\\hline
$h$  & P & R & F1 & AUC-ROC & P & R   & F1 & AUC-ROC  & P & R   & F1 & AUC-ROC   \\\hline
8   & 83.13  & 62.75  & 71.52  & 81.33    & 41.56  & 46.83 & 44.04 & 65.67 & 58.13  & 43.11 & 49.51 & 65.99  \\
16  & 83.59  & 62.86  & 71.76  & 81.72    & 42.31  & 46.97 & 44.52 & 65.75 & 59.20  & 44.94 & 51.09 & 68.92 \\
32  & 84.98  & 63.54  & 72.71  & 82.39    & 42.14  & 46.61 & 44.26 & 65.70 & 59.31  & 44.07 & 50.57 & 66.91  \\
64  & 87.67 & 61.36 & 72.19  & 82.33  
   & 42.59  & 45.10 & 43.81 & 65.18 & 59.85  & 44.03 & 50.73 & 68.57  \\
 \hline
\end{tabular}}
\end{table*}

\begin{table*}
\centering
\caption{Ablation study. }
\label{table:ablation}
\resizebox{0.92\textwidth}{!}
{
\begin{tabular}{c|ccc|ccc|ccc|ccc} \toprule
\multicolumn{1}{c|}{Dataset} & \multicolumn{3}{c|}{PSM} & \multicolumn{3}{c|}{SWAN} & \multicolumn{3}{c|}{SWaT} & \multicolumn{3}{c}{GECCO} \\ \hline
 Metric & P & R & F1 & P & R & F1 & P & R & F1 & P & R & F1 \\ \hline\hline
w/o APPGM & 41.58 & 40.51 & 41.04 & 55.82 & 41.03 & 47.30 & 84.19 & 60.28 & 70.26 & 51.88 & 29.03 & 37.23 \\
w/o regularization loss & 41.39 & 43.70 & 42.51 & 57.14 & 41.81 & 48.29 & 84.45 & 63.01 & 72.17 & 52.03 & 31.90 & 39.55  \\
w/o memory bank & 41.42 & 42.86 & 42.13 & 56.24 & 41.50 & 47.76 & 84.59 & 61.25 & 71.05 & 52.27 & 31.24 & 39.11 \\
w Transformer & 41.07 & 44.23 & 42.59 & 57.08 & 42.63 & 48.80 & 84.26 & 62.86 & 72.00 & 52.20 & 32.12 & 39.77 \\
\textbf{IGCL}  & 42.31 & 46.97 & \textbf{44.52} & 59.20 & 44.94 & \textbf{51.09} & 84.98 & 63.54 & \textbf{72.71} & 52.10 & 32.77 & \textbf{40.23} \\ \bottomrule
\end{tabular}
}
\end{table*}

We evaluate the anomaly prediction with larger look-forward windows $f$ in Table~\ref{tab:Largerlook-forward}. We can see that IGCL still performs better than state-of-the-art baselines, even though the farther future anomalies are more difficult to predict.

The impact of the size of look-back windows $h$ is in Table~\ref{tab:look-backwindow}. We can see that IGCL is relatively stable to $h$ and time series from different domains require different look-back windows to better identify the anomaly precursors.

\subsection{Ablation Studies} 
\label{subsec:Ablation}

We conduct ablation studies to validate the
effectiveness of our key components.
In particular, we compare IGCL with the following variants:
\noindent\begin{itemize}[leftmargin=*]
\item w/o anomaly precursor pattern generation module (APPGM): 
This variant directly uses random Gaussian noises as anomaly precursor patterns without the diffusion process. 
\item w/o regularization loss: This variant does not have regularization loss to control the diffusion process. 
\item w/o memory bank: This variant does not use the memory bank, and thus precursor patterns are generated during each iteration and appear in individual variable. 
\item w Transformer: This variant does not use the overlapping window-based TCN. It uses naive Transformer~\cite{attention} to learn representations across timestamps.
\end{itemize}

Table~\ref{table:ablation} shows the accuracy of different variants. From Table~\ref{table:ablation} we
observe that: (1) As our anomaly precursor pattern generation module, regularization loss, and memory bank all help generate more complicated and representative anomaly precursors as negative samples, our model can have more accurate results. 
(2) Our overlapping window-based TCN can effectively learn the temporal dependencies from pairs of successive sub-sequences, and it is also more efficient than naive Transformer as shown in the complexity analysis.

\begin{figure}[t]
\centering 
\includegraphics[width=0.48\textwidth]{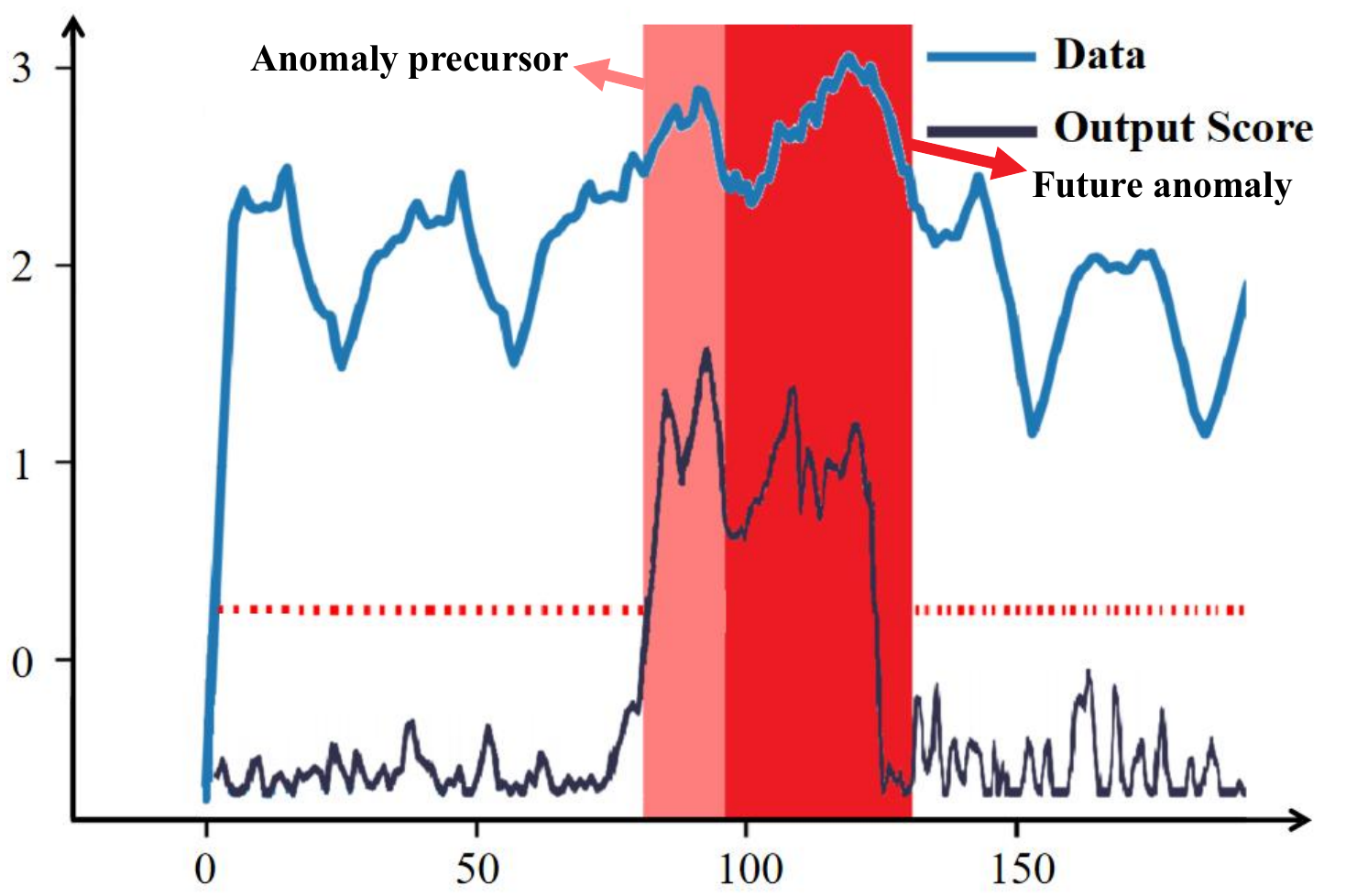}
\caption{IGCL can output larger scores on the precursors that start to deviate from the normal ahead of the severer anomalies.}
\label{fig:Visualization}
\end{figure}

\subsection{Case Study and Visualization}
We visualize the anomaly prediction of our IGCL method on the SMAP dataset, as shown in Figure~\ref{fig:Visualization}. By identifying the anomaly precursor that just begins to deviate from the normal, our model can report a warning at timestamp 80, ahead of the severe anomalies after timestamp 100.

\begin{table*}[t]
\caption{Comparisons between representative methods that can be applied to unsupervised time series anomaly prediction.}
\label{tab:baselineComparisons}
\resizebox{0.82\textwidth}{!}
{  
\begin{tabular}{c | c | c | c | c} \hline
Methods  & {For time series}  & {Unsupervised}  & {Precursor Generation} & \makecell[c]{Precursor Dependencies} \\ \hline
~\cite{DeepOneClass, LOF, KmeanIsolationForest}        & \xmark  & \cmark  & \xmark & \xmark\\ \hline
~\cite{Earthquake, EarlyAnomaly, EarlyAnomalynew}     & \cmark & \xmark  & \xmark  & \xmark \\ \hline
~\cite{outpre1, outpre2, outpre00, outpre0}     & \cmark & \xmark  & \xmark  & \cmark \\ \hline
\makecell[c]{~\cite{ImDiffusion, CAE-Ensemble, outreview, ITAD, THOC, Autoencoder, Autoencoder1, Autoencoder2, SMD-omni, InterFusion, GANomaly} \\ ~\cite{VQRAE, PUAD, Precursor-of-Anomaly,  DiffAD, D3R, DC, AnomalyTrans, DAGMM} }     & \cmark  & \cmark  & \xmark & \xmark\\   \hline
Our IGCL  & \cmark   & \cmark & \cmark & \cmark  \\   \hline
\end{tabular}
}
\end{table*}
\section{Related Works}
\label{sec:related}

We review the relevant works on time series anomaly prediction, anomaly detection, contrastive learning, and generative learning. We also compare our method with related works in Table~\ref{tab:baselineComparisons}.

\subsection{Time Series Anomaly Prediction}
Based on the current time series data, anomaly prediction aims to make the real-time judgment on whether there are starting anomaly precursors where data begin to deviate from normal to future anomalies. Anomaly prediction plays an important role in many scenarios that prioritize the safety or economic influence of anomalies, However, there are a few studies~\cite{EarlyAnomaly, EarlyAnomalynew, Earthquake, Precursor-of-Anomaly, outpre0, outpre00, outpre2, outpre1} related to this problem. 
FEP~\cite{Earthquake} extracts statistical information from manually labeled frequency-based features as the starting anomaly precursors to predict future anomalies.   
EADS~\cite{EarlyAnomaly, EarlyAnomalynew} and VFP~\cite{outpre0} use graphs to extract correlations among variables to help classify the starting anomaly precursors related to different variables.
MCDA~\cite{outpre1} proposes multi-instance attention, dCNN~\cite{outpre00} proposes deep convolutional neural networks and CHE~\cite{outpre2} proposes hierarchical windows, to extract temporal dependency features.
They learn temporal dependencies from the pair of the normal time series sub-sequences and the starting anomaly precursors which help better identify starting abnormal changes for anomaly prediction.

However, these existing methods heavily rely on supervised learning of the temporal dependencies to get accurate anomaly prediction results. They are limited and fall short in the unsupervised time series anomaly prediction task for practical applications. 

\subsection{Time Series Anomaly Detection}
There has been a lot of research on time series anomaly detection.  Early works~\cite{anomlyreview} utilize classification-based methods to detect anomalies in the historical time series data.
Recent methods mainly focus on unsupervised settings based on the one-class classification theory~\cite{DeepOneClass}. 
They can be further categorized into four categories~\cite{outyin2}, including density estimation-based methods~\cite{LOF}, clustering-based methods~\cite{outyin2}, autoregression and reconstruction-based methods~\cite{SMD-omni} methods, and contrastive-based methods~\cite{DC}.  

The clustering-based methods, such as K-means~\cite{outreview}, Deep-SVDD~\cite{DeepOneClass}, and ITAD~\cite{ITAD}, aim to cluster normal samples within a center. THOC~\cite{THOC} uses a hierarchical RNN to learn dynamic clustering centers for time series data.

The density estimation-based anomaly detection methods, such as LOF~\cite{LOF}, assume abnormal samples lie in a low-density region. IForest~\cite{KmeanIsolationForest} uses the tree model and DAGMM~\cite{DAGMM} uses the encoding probability model to measure the density or distribution.  
LSTM~\cite{outreview} proposes forecasting-based approaches that forecast values and based on the forecasting errors detect whether the time points are anomalies. 
Autoencoder~\cite{Autoencoder, Autoencoder1, Autoencoder2} (AE) is a classic architecture used in reconstruction. Omni~\cite{SMD-omni}, InterFusion~\cite{InterFusion} and LSTM-VAE~\cite{VQRAE} utilize the Variational AE and use reconstruction probability to detect anomalies. GANomaly~\cite{GANomaly} introduces a discriminator to generate normal data and detect anomalies based on reconstruction errors.  CAE-Ensemble~\cite{CAE-Ensemble} proposes to ensemble different AEs for reconstruction.
PUAD~\cite{PUAD} uses meta-learning-oriented AE to reconstruct typical time series.

\textbf{Generative Learning.}
The Generative Adversarial Network (GAN) is one kind of generative model, composed of a generator and a discriminator~\cite{DBLP:journals/corr/vae}. By designing the minimax theoretical game, the generator and discriminator improve each other and learn the underlying data distribution in an unsupervised setting. 
The diffusion-based network is a generative model that generates data with denoising steps~\cite{DiffAD}. ImDiffusion~\cite{ImDiffusion}, DiffAD~\cite{DiffAD} and D3R~\cite{D3R} utilize diffusion-based networks to generate normal time series data and detect anomalies based on reconstruction errors. However, these methods only learn the underlying normal data distribution, but cannot generate anomaly precursors.

\textbf{Contrastive Learning.} Contrastive learning learns useful representations by contrasting positive pairs against negative pairs with data augmentation~\cite{SimCLR}. 
Recently, contrastive learning has been proposed for anomaly detection, such as DCdetector~\cite{DC}, where two view representations of a normal timestamp should be similar, and contrastive errors are used to detect anomalies. ATransformer~\cite{AnomalyTrans} combines the reconstruction errors and contrastive errors together to detect anomalies.

PAD~\cite{Precursor-of-Anomaly} proposes to utilize unsupervised anomaly detection methods and adjust specific thresholds on the anomaly scores to identify the anomaly precursors for anomaly prediction. However, these methods do not have negative samples and only reconstruct or contrast positive samples. They cannot learn the temporal precursor dependencies from the pair of normal time series and the anomaly precursor. Their performance is limited compared to supervised learning with labeled anomaly precursors as negative samples~\cite{outpre2, outpre1, outpre00, outpre0}.
\section{CONCLUSION}
\label{sec:conclusion}

We present an importance-based generative contrastive learning model for a novel problem of unsupervised time series anomaly prediction. We propose anomaly precursor pattern generation, with diffusion-based transformation and variance regularization, to generate diverse anomaly precursors as labeled negative data. Then, we propose an overlapping window-based contrastive learning to distinguish anomaly precursors from the normal efficiently. Last, we propose a memory bank with importance scores to store representative anomaly precursors to generate complicated precursors efficiently. Experiments on seven benchmark datasets demonstrate the superiority of our method. In future works, it is worth improving IGCL with pre-training on large-scale time series data.

\begin{acks}
This work was partially supported by National Natural Science Foundation of China (62372179), Independent Research Fund Denmark (8022-00246B), and the VILLUM FONDEN (40567).
\end{acks}

\bibliographystyle{ACM-Reference-Format}
\balance
\bibliography{sample-base}

\newpage

\newpage

\appendix

\section{Proof}
\label{Appendix:Proof}

\noindent \textbf{Theorem 1: } Under the unsupervised setting, the future time series anomaly is unpredictable if we cannot identify the anomaly precursor in the current observation that follows the same distribution as the normal previous sub-sequences do.

\noindent \textbf{Proof 1: }
As the current observations follow the same conditional distribution as the normal sub-sequence follows, we have $\mathcal{N} = \mathcal{A}$. For any model $\mathcal{F}$ we have:
\begin{equation}
\label{E:samedistribution}
\begin{aligned}
\mathop{P}\limits_{\mathbf{H_t} \sim \mathcal{A}} \big [ \mathcal{F}(\mathbb{X}^b_{t, h}) = \hat{p}_{t} \big ] - \mathop{P}\limits_{\mathbf{H_t} \sim \mathcal{N}} \big [ \mathcal{F}(\mathbb{X}^b_{t, h}) = \hat{p}_{t}   \big ] = 0.
\end{aligned}
\end{equation}
Thus, the predicted probability score for the anomaly precursor in the current observation is equal to that for the normal sub-sequences. Therefore, no unsupervised model can predict the future anomaly if we cannot identify the anomaly precursors in the current observations when the anomaly precursor follows the same distribution as the normal sub-sequences do. $\square$

\noindent \textbf{Theorem 2: } Under the unsupervised setting, the future time series anomaly is predictable if we can identify its anomaly precursor in the current observation that follow a distribution different from the normal sub-sequences.

\noindent \textbf{Proof 2: }
As the anomaly precursor follows a data distribution different from the normal sub-sequences, we have $\mathcal{N} \neq \mathcal{A}$. According to the Maximum Mean Discrepancy (MMD) theory~\cite{MMD}, there exists at least one mapping function $\mathcal{F}$ satisfying:
\begin{equation}
\label{E:differentdistribution}
\begin{aligned}
\mathop{\mathbb{E}}\limits_{\mathbf{H_t}\sim \mathcal{A}} \big[\mathcal{F}(\mathbb{X}^b_{t, h}) \big] - \mathop{\mathbb{E}}\limits_{\mathbf{H_t}\sim \mathcal{N}} \big[\mathcal{F}(\mathbb{X}^b_{t, h}) \big]  > 0
\end{aligned}
\end{equation}
Thus, the expectation of the predicted probability score for the anomaly precursor is larger than that for the normal sub-sequence. Therefore, there exists at least one model $\mathcal{F}$ that can predict a larger probability score to identify that there is the starting anomaly precursor and the future time series is more like an anomaly when the anomaly precursor follows the data distribution different from the normal sub-sequences. $\square$

Based on the above analysis, we can only predict the future time series anomalies whose anomaly precursors just begin to deviate from the nearest previous normal sub-sequences under the unsupervised setting. However, we do not have labeled anomaly precursor data for the model to learn the temporal dependencies to estimate the Markov conditional distributions for $\mathcal{A} = P (\cdot | \mathbf{H_{t-h-1}}, \mathbf{H_t} \text{ is anomaly}$  $\text{precursor})$ and $\mathcal{N} = P (\cdot | \mathbf{H_{t-h-1}}, \mathbf{H_t}$ is normal$)$. Thus, the best potential model is that we generate different kinds of potential anomaly precursors to simulate the beginning of deviation from normal and then learn the temporal dependency features. 

\section{The pollution process}
\label{Appendix:noise pollution}

The noise pollution process $q(x^s|x^{s-1})$ is shown as
\begin{equation}
\label{eq:pollution}
\begin{aligned}
    x^s = \sqrt{1-\beta^s} x^{s-1} + \sqrt{\beta^s} \epsilon^s, \  \forall s\in \{1, 2, \cdots, S \},
\end{aligned}
\end{equation}
where $\epsilon^s \sim {N}(0,I)$ are the independent Gaussian pollutions to be added in different steps, $I$ is the identity matrix, $0 < \beta^s < 1$ are the hyper-parameters control the amount of pollution added in each step.
Thus, we have: 
\begin{equation}
\label{eq:pollution1}
\begin{aligned}
x^s &= \sqrt{1-\beta^s} x^{s-1} + \sqrt{\beta^s} \epsilon^s \\
\\\end{aligned}
\end{equation} 

$= \sqrt{1-\beta^s} (\sqrt{1-\beta^{s-1}} x^{s-2} + \sqrt{\beta^{s-1}} \epsilon^{s-1}) + \sqrt{\beta^s} \epsilon^s  $

$ = \cdots$

$ = (\sqrt{1-\beta^s} \sqrt{1-\beta^{s-1}} \cdots \sqrt{1-\beta^1}) x^0 \, + $

$ \, \, \, \, \, \, \, (\sqrt{1-\beta^s} \sqrt{1-\beta^{s-1}} \cdots \sqrt{1-\beta^2})\sqrt{\beta^1} \epsilon^1 \, +  $

$ \, \, \, \, \, \, \, (\sqrt{1-\beta^s} \sqrt{1-\beta^{s-1}} \cdots \sqrt{1-\beta^3})\sqrt{\beta^2} \epsilon^2 \, + \cdots \, + $

$ \, \, \, \, \, \, \, \sqrt{1-\beta^s} \sqrt{\beta^{s-1}} \epsilon^{s-1} + \sqrt{\beta^s} \epsilon^s$

\noindent For any independent Gaussian distributions $\epsilon_1 \sim {N}(\mu_1,\sigma^2_1)$ and $\epsilon_2 \sim {N}(\mu_2,\sigma^2_2)$, we have the following properties:
\begin{equation}
\begin{aligned}
a\epsilon_1+b  \sim {N}(a\mu_1+b,a^2\sigma_1^2),
\end{aligned}
\end{equation}
\begin{equation}
\begin{aligned}
 \epsilon_1+ \epsilon_2 \sim {N}(\mu_1+\mu_2,\sigma^2_1+\sigma^2_2).
\end{aligned}
\end{equation}
We also have:
\begin{equation}
\begin{aligned}
  1 &= \big ( \sqrt{1-\beta^s} \sqrt{1-\beta^{s-1}} \cdots \sqrt{1-\beta^1} \big )^2 +\\
&\, \, \, \, \, \, \, \big ( \sqrt{1-\beta^s} \sqrt{1-\beta^{s-1}} \cdots \sqrt{1-\beta^2} \big )^2 {\beta^1}  \, +  \\
&\, \, \, \, \, \, \, \big (\sqrt{1-\beta^s} \sqrt{1-\beta^{s-1}} \cdots \sqrt{1-\beta^3} \big )^2 {\beta^2} \, + \cdots \, + \\
&\, \, \, \, \, \, \, ({1-\beta^s}){\beta^{s-1}} + {\beta^s},
\end{aligned}
\end{equation}
Thus, we have:
\begin{equation}
\label{eq:pollution2}
\begin{aligned}
    x^s = \sqrt{\widetilde{\alpha}^s} x^0 + \sqrt{1-\widetilde{\alpha}^s} \epsilon,
\end{aligned}
\end{equation}
where $\epsilon \sim {N}(0,I)$ and $\sqrt{\widetilde{\alpha}^s}= \sqrt{1-\beta^s} \sqrt{1-\beta^{s-1}} \cdots \sqrt{1-\beta^1}$. Then, for enough $S$ steps and proper $0 < \beta^s < 1$,  $\sqrt{\widetilde{\alpha}^s}$ will be close to $0$, and the output from the noise pollution process, \ie $x^S$, will be gradually corrupted into random Gaussian distribution.

\begin{table*}
\centering
\caption{More experiment results of state-of-the-art methods, presented in percentages.}
\label{table:AUCperformance}
\resizebox{0.99\textwidth}{!}
{
\begin{tabular}{c|c|c|c|c|c|c|c|cccc|cccc} \hline
Dataset & \multicolumn{1}{c|}{PSM} & \multicolumn{1}{c|}{SMAP} & \multicolumn{1}{c|}{SWAN} & \multicolumn{1}{c|}{SWaT} & \multicolumn{1}{c|}{GECCO} & SMD & MSL & \multicolumn{4}{c|}{UCR}& \multicolumn{4}{c}{Credit}\\ \hline
Metrics & \multicolumn{7}{c|}{AUC-ROC} & P & R & F1 & AUC-ROC & P & R & F1 & AUC-ROC\\ \hline\hline
DCdetector   & 50.91 & 43.98 & 51.00 & 65.57 & 50.79 & 50.44 & 30.34 & 12.59 & 10.90 & 11.68 & 48.65 & 37.50 & 12.96 & 19.26 & 64.34\\
PAD          & 51.02 & 50.13 & 50.85 & 61.35 & 49.71 & 49.66 & 48.28 & 29.92 & 20.35 & 24.22 & 51.67 & 31.49 & 16.01 & 21.23 & 68.76 \\\hline
ATransformer & 54.91 & 53.70 & 51.95 & 75.63 & 50.39 & 59.30 & 50.21 & 32.07 & 20.86 & 25.28 & 51.68 & 17.58 & 26.85 & 21.25 & 68.55\\
CAE-Ensemble & 63.68 & 58.76 & 65.41 & \underline{81.71} & 52.94 & \underline{67.99} & {51.26} & {34.15}  & {21.47} & 26.37 & 58.26 & 18.16 & 45.98 & 26.04 & \underline{82.16} \\ 
D3R & \underline{63.93} & \underline{59.61} & \underline{66.13} & 80.62 & \underline{53.09} & 66.83 & \underline{51.83} & 39.10 & 20.77 & \underline{27.13}  & \underline{60.11}&  17.85 &  50.17 &  \underline{26.33} &  80.68\\ \hline
\textbf{IGCL} & \textbf{65.75} & \textbf{61.92} & \textbf{68.92} & \textbf{82.39} & \textbf{54.02}  & \textbf{69.97}  & \textbf{55.89} & {36.43} & {31.04}& \textbf{33.52} & \textbf{69.82} & {17.62} & 59.07 & \textbf{27.15} & \textbf{84.00} \\ \hline
\end{tabular}
}
\end{table*}

\begin{table*}[t]
\caption{Anomaly prediction for irregular time series}
\label{tab:Irregular}
{
\begin{tabular}{c|cccc|cccc|cccc}\hline
Dataset               & \multicolumn{12}{c}{SWaT} \\\hline
Dropping ratio   & \multicolumn{4}{c|}{0\%}    & \multicolumn{4}{c|}{30\%}  & \multicolumn{4}{c}{50\%} \\\hline
Metric  & P & R & F1 & AUC & P & R & F1 & AUC & P & R &F1 & AUC  \\\hline
PAD & 74.83 & 59.09 & 66.04 &  61.35  & 74.17 & 57.80 & 64.97 & 60.71  & 69.29 & 50.94 & 58.71 & 56.09\\
IGCL+ & 84.98 & 63.54 & \textbf{72.71} & \textbf{82.39} & 84.02 & 59.97 & \textbf{69.98} & \textbf{78.63} & 70.81 &  53.77 & \textbf{61.13} & \textbf{ 60.17
} \\\hline
\end{tabular}
}
\end{table*}

\section{Regularization loss}
\label{Appendix:Regularization}

\begin{equation}
\begin{aligned}
&KL\Big({N}(\mu,{\sigma}^2)\Big\Vert N(\mu,I)\Big)\\ 
=&\int \frac{1}{\sqrt{2\pi\sigma^2}}e^{-(x-\mu)^2/2\sigma^2} \left(\log \frac{e^{-(x-\mu)^2/2\sigma^2}/\sqrt{2\pi\sigma^2}}{e^{-(x-\mu)^2/2}/\sqrt{2\pi}}\right)dx\\ 
=&\int \frac{1}{\sqrt{2\pi\sigma^2}}e^{-(x-\mu)^2/2\sigma^2} \log \left\{\frac{1}{\sqrt{\sigma^2}}\exp\left\{\frac{1}{2}(x-\mu)^2(1-\frac{1}{\sigma^2})\right\} \right\}dx\\
=&\frac{1}{2}\int \frac{1}{\sqrt{2\pi\sigma^2}}e^{-(x-\mu)^2/2\sigma^2} \Big[-\log \sigma^2+(x-\mu)^2(1-\frac{1}{\sigma^2}) \Big] dx\\ 
=&\frac{1}{2}\Big(-\log \sigma^2+\sigma^2(1-\frac{1}{\sigma^2})\Big)\\
=&\frac{1}{2}\Big(-\log \sigma^2+\sigma^2-1\Big).\\
\end{aligned}
\end{equation}

\section{Experimental details}
\label{Appendix:Experiment}

\subsection{Datasets}
\label{Appendix:Datasets}

\begin{itemize}[leftmargin=3.5mm]
\item SMD~\cite{SMD-omni} (Server Machine Dataset) is collected from a large computing cluster, where different server machines stack accessed traces of resource utilization with 38 dimensions. 
\item PSM~\cite{PSM} (Pooled Server Metrics) is a 25-dimension dataset collected by eBay that shows the performance of multiple application servers.
\item MSL~\cite{SMAP-MSL} (Mars Science Laboratory Rover) is from the spacecraft monitoring systems and collected by NASA. It shows the health check-up data of the sensors from the Mars rover. 
\item SMAP~\cite{SMAP-MSL} (Soil Moisture Active Passive) is a 25-dimension dataset collected by NASA, which contains soil samples and telemetry information. 
\item SWAN~\cite{NIPS-DS} is extracted from solar photospheric vector magnetograms in the Spaceweather, which is the benchmark dataset used in NeurIPS Competition Track.
\item SWaT~\cite{AnomalyTrans} (Secure Water Treatment) is collected from sensors of the critical infrastructure systems under continuous operations for secure water treatment.
\item GECCO~\cite{NIPS-DS} is a 9-dimension dataset collected from the cyber-physical systems showing the drinking water quality.  
\item UCR~\cite{outreview2} is a 1-dimension dataset containing 250 sub-datasets from various domains.
\item Credit~\cite{anomlyreview} is from the finance scenario where the anomaly rate is only 0.172\%.
\end{itemize}

These datasets are representative open benchmarks for multi-variable and uni-variable time series anomaly tasks, and thus we evaluate the unsupervised time series anomaly prediction problem on these datasets. More statistical information and details can be seen in Table~\ref{table:BenchmarkDatasets}, where $N$ is the number of variables observed at each timestamp. We follow the same training-validation-test splits as in the original papers~\cite{SMD-omni, PSM, SMAP-MSL, AnomalyTrans, NIPS-DS, CAE-Ensemble} shown in each column.

\subsection{Settings}
\label{Appendix:Settings}
We employ the official open-source implementations of baseline methods and have carefully tuned their hyper-parameters based on the recommendations from the original papers. The target label from the look-forward window is modified for the anomaly prediction task following PAD~\cite{Precursor-of-Anomaly}. The anomaly score from the current look-back window is used for identifying the anomaly precursors to predict future anomalies also following PAD~\cite{Precursor-of-Anomaly}. The experiments are conducted on an Ubuntu 18.04.5 LTS system server, with Intel(R) Xeon(R) Gold 5215 CPU @ 2.50GHz and NVIDIA Quadro RTX 8000 GPU. All the deep learning-based models are executed with Pytorch 1.2.0. 
The size of the default look-forward window size $f$ is 4, and we also evaluate with larger look-forward windows of \{4, 8, 12\}. We use Adam optimizer with the default parameter and the learning rate is default 0.0001. The batch size is tuned from \{32, 64\}. The look-back window size $h$ and the number of sub-sequences $b$ are tuned from \{8, 16, 32, 64\}. The memory bank size $K$ is tuned from \{16, 20, 24\} and we evaluate with more sizes in the parameter study. The number of positive samples $P$ is tuned from \{12, 16, 20\}. The $\lambda$ which controls the strength of regularization is tuned from \{0.5, 1\}. We use three parallel kernel sizes \{2, 3, 5\} for our overlapping TCN. Following existing works~\cite{anomlyreview, SMD-omni}, we choose the best threshold for all methods.

\subsection{More experiment results}
\label{Appendix:More experimental results}
Table~\ref{table:AUCperformance} shows more experiment results of most state-of-the-art methods on all datasets using more metrics.

\subsection{Irregular time series}
PAD uses interpolation and neural controlled differential equations (NCDE) to support irregular time series. The interpolation method can be applied before any models. Our TCN backbone can also be replaced with NCDE. We show the experimental results on the irregular series by dropping random points following PAD. We only add the interpolation method before our model, which is denoted as IGCL+. We can see that our method can be extended to irregular series easily and still perform well.

\end{document}